\documentclass[conference]{IEEEtran}
\IEEEoverridecommandlockouts

\usepackage{cite}
\usepackage{amsmath,amssymb}
\usepackage{graphicx}
\usepackage{booktabs}
\usepackage{array}
\usepackage{url}
\usepackage{pgfplots}
\usepackage{enumitem}
\usetikzlibrary{patterns}
\pgfplotsset{compat=1.18}

\begin{document}
\raggedbottom 

\title{BoardSafety: AI-Ready Data from Manufacturer\\
         Documentation for Development-Board Safety}

\author{
\IEEEauthorblockN{Linjian Yang}
\IEEEauthorblockA{\textit{Clemson University}\\
Clemson, SC, USA\\
linjiay@clemson.edu}
\and
\IEEEauthorblockN{Xinyan Wang}
\IEEEauthorblockA{\textit{Portland State University}\\
Portland, OR, USA\\
xinyw@pdx.edu}
\and
\IEEEauthorblockN{Kunpeng Liu}
\IEEEauthorblockA{\textit{Clemson University}\\
Clemson, SC, USA\\
kunpenl@clemson.edu}
}

\maketitle

\begin{abstract}

The development of large language models (LLMs) are capable to assist with hardware experiments. In this scenario, this raises a practical question: Can LLMs determine whether a proposed operation is safe for a specific development board? Documented operating limits and safety requirements are board-specific, even across products from the same manufacturer. This variation makes board-specific safety judgments by an LLM difficult to evaluate. 
To address this problem, we introduce the \textbf{BoardSafety Benchmark}. The benchmark is built from vendor datasheets and hardware guides. For data transformation, these documents are converted into a board-specific safety rule base and are then used to create labeled operating scenarios. Each data item pairs a board and a proposed operation to an applicable safety rule with ground-truth label, indicating the level of safety. The categorized level for each operation includes \texttt{safe}, \texttt{unsafe}, or \texttt{insufficient-info}. 
In addition, the benchmark covers diverse operations, board-specific cases, boundary conditions, and missing-information cases. 
We evaluate a series of models by varying their access to board-specific rules and the use of structured-rationale prompting.
Our results demonstrate that providing the safety rules generally improves accuracy, whereas structured-rationale prompting produces gains that vary across models and rule-access conditions. Overall, the evaluated models cannot yet reliably determine whether an operation is safe, which implies that critical exploration is required for this field.

\end{abstract}


\section{Introduction}

\begin{figure*}[t]
    \centering
    \includegraphics[width=\textwidth]{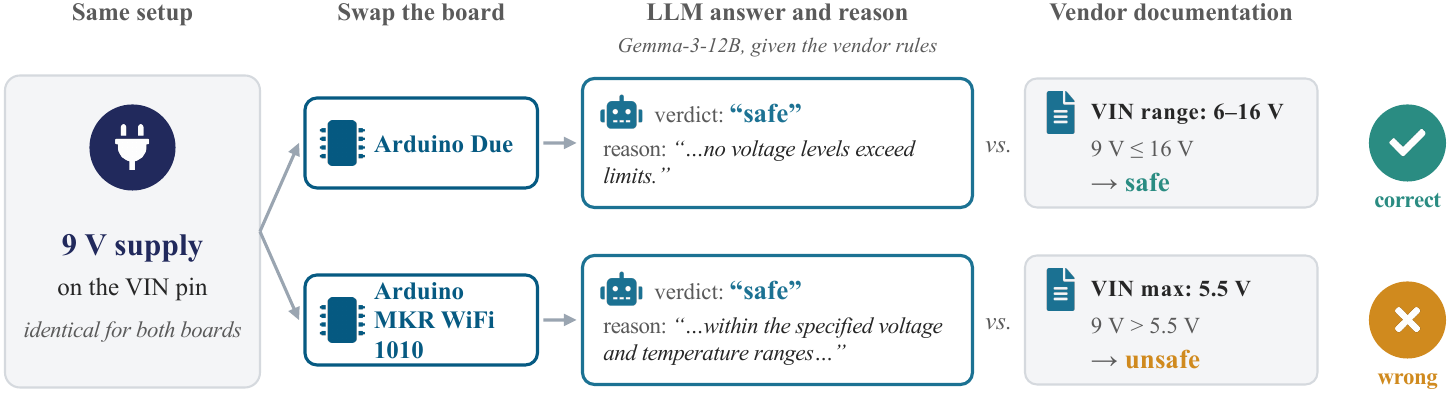}
    \caption{Despite access to manufacturer rules, Gemma-3-12B gives the same safety verdict to a board-swap pair with different VIN limits.}
    \label{fig:running-example}
    \vspace{-4mm}
\end{figure*}

Large language models (LLMs) are increasingly consulted during embedded system development and debugging~\cite{englhardt2024exploring}. Unlike a software bug, an LLM's response that leads the hardware to go beyond its operational limits can permanently damage the board and introduce significant safety risks. Assessing whether an operation is safe requires more than general electronics knowledge, because the governing limits are specific to each board and differ even between products of the same manufacturer. Therefore, an LLM must identify the applicable manufacturer-specified constraint for the given board and setup. Figure~\ref{fig:running-example} shows two boards connected to the same 9~V supply through their VIN pins. This voltage is within the documented 6--16~V range for the Arduino Due~\cite{arduino_due_ds}, but exceeds the 5.5~V limit for the Arduino MKR WiFi 1010~\cite{arduino_mkr1010_ds}. The correct verdict is therefore \texttt{safe} for the Due and \texttt{unsafe} for the MKR WiFi 1010 under the stated constraints. However, Gemma-3-12B labels both operations \texttt{safe} and claims that neither violates the voltage limits, even though both boards' rules are provided in its context. This example motivates our research question: Can LLMs reliably determine whether a proposed operation is safe for a specific development board?

Prior studies show that LLM advice can pose safety risks in specific contexts~\cite{levy2022safetext, wu2025first}, and that in specialized domains, a safe response requires the model to retrieve relevant domain constraints and apply them to the specific situation~\cite{hager2024evaluation, zhou2024labsafety}. Hardware benchmarks offer a complementary perspective by grounding LLM evaluation in concrete engineering artifacts and verifiable criteria. These benchmarks~\cite{skelic2025circuit, srinivasan2026pcb, li2026pcb, qiu2026hwe, xu2026domusmind, xu2026embedagent} assess whether a generated design or implementation is valid for the target platform, but none test whether an LLM selects and applies the manufacturer constraint governing a proposed operation on a specific development board. We address this gap by evaluating LLM safety judgments against the documented constraints manufacturers provide for the specified board.

Although manufacturer documentation is the primary source of hardware-safety constraints, it is not directly usable for evaluating LLM judgments. Safety-relevant information is scattered across datasheets, hardware guides, and reference manuals, mixing hard operating limits with recommendations and descriptive specifications~\cite{chen2024doceda}. These limits can take different forms, including numeric ranges, maximum values, prohibited configurations, and procedural notes. Making such documentation AI-ready therefore means more than collecting the information~\cite{wilkinson2016fair,zhang2026braillebench,ma2023comprehensive}. Each constraint must be traceable to its source, encoded so a proposed operation can be compared mathematically against it, and tagged to indicate how much information is directly documented rather than inferred. We introduce \textbf{BoardSafety}, which implements these requirements as five properties of its rule base: traceable, computable, filterable, consistent, and derivable.

BoardSafety consists of two artifacts. The first is a rule base containing 490 constraint records extracted from 30 manufacturer documents that cover 53 development boards from five vendors. Each record in this rule base includes a normalized constraint, the document source, and an evidence tier. The second is a benchmark of 298 scenarios derived from this rule base. Each scenario pairs a specific board and a proposed operation with the applicable constraint, and a gold label of \texttt{safe}, \texttt{unsafe}, or \texttt{insufficient-info}. We fix each label by comparing the stated operating value or configuration with the documented constraint before an LLM renders the scenario text so that we can recompute every label from the rule base alone. The scenarios include minimal pairs that vary only the board or only the decisive value, as well as underspecified cases that omit the decisive value. We use the benchmark to study how rule access and structured-rationale prompting affect the LLM judgments.

Our experiments on six open-weight LLMs yield a maximum accuracy of 69.13\% across 7,152 responses, with only four of 24 model--condition combinations exceeding the 48.3\% always-unsafe baseline by their 95\% Wilson lower bounds. Access to rules yields larger and more consistent gains than structured-rationale prompting, yet only 13.4\% of 216 rationales in a stratified review correctly identify and apply the applicable constraint. Our findings highlight persistent limitations in constraint-based reasoning, particularly in board-swap, near-boundary, and underspecified cases.

\smallskip
\noindent This paper makes the following contributions:

\begin{itemize}

    \item We build an AI-ready rule base of 490 manufacturer-documented constraints with five verifiable properties, so every gold label is traceable and recomputable.

    \item We construct the BoardSafety benchmark for board-specific hardware safety judgment, testing whether LLMs apply board-specific constraints and abstain when information is missing.

    \item Our experiments on six LLMs show that aggregate accuracy masks persistent failures in board sensitivity, boundary handling, abstention, and constraint application.

\end{itemize}

\section{Related Work} 

\subsection{Context-Specific Safety of LLM Responses}


LLMs are expected to generate advice that is safe and appropriate for the context in which it is used. Existing studies examine harmful advice in everyday situations and safety failures in specialized domains~\cite{levy2022safetext, hager2024evaluation,zhang2025ratt, zhou2024labsafety}. However, beyond recognizing hazards, models must apply the relevant constraints to each case. They should also abstain when the available information is insufficient~\cite{madhusudhan2025llms,mo2026opendecoder, xu2026shift, hao2025generative}. For example, judging a power connection requires comparing the supply voltage with the board's input limits. A model should request this safety information before giving a safety verdict.

Existing studies evaluate LLM safety through commonsense advice, clinical decision-making, and laboratory risk assessment. Hager et al.~\cite{hager2024evaluation} found that LLMs struggled to use relevant patient information and follow clinical guidelines in realistic decision-making tasks. NOHARM~\cite{wu2025first} shows that LLMs performing strongly on medical knowledge benchmarks can still generate potentially harmful recommendations. LabSafety Bench~\cite{zhou2024labsafety} reports similar limitations in hazard identification and risk assessment for laboratory procedures. These studies show that safe responses require models to apply domain-specific evidence to the specific decision context.

LLMs may ground their judgments on both parametric knowledge and contextual information provided in the prompt~\cite{xu2024knowledge, xie2024adaptive, yang2026thoughts}, but they do not reliably reconcile these sources when they conflict~\cite{xie2024adaptive, wang2023resolving}. A verdict alone therefore does not reveal which source grounds it. The BoardSafety examines this problem in development-board operations by evaluating whether model responses follow the relevant manufacturer constraint.
 

\subsection{LLM Benchmarks for Hardware and Embedded Systems}
Hardware benchmarks evaluate LLMs on circuit understanding and hardware design tasks. These tasks require specialized engineering knowledge and physical constraints that general language and coding benchmarks do not adequately assess. Existing benchmarks broadly cover two areas: interpreting existing circuits and generating hardware designs. Most focus on engineering reasoning and design correctness. For circuit interpretation, CIRCUIT tests analog-circuit understanding and electrical reasoning~\cite{skelic2025circuit}, while PCB-QA and PCB-Bench examine connectivity, datasheet information, placement, routing, and design-rule compliance~\cite{srinivasan2026pcb, li2026pcb}. For design generation, HWE-Bench evaluates board-level schematics through electrical-rule checks and circuit simulation~\cite{qiu2026hwe}. However, these benchmarks do not specifically assess whether a proposed development-board operation complies with manufacturer-documented safety limits.

Embedded-system evaluations focus on implementation and execution in hardware-dependent settings. Englhardt et al. evaluate generated embedded software using an automated hardware-in-the-loop testbench~\cite{englhardt2024exploring}. EmbedBench covers embedded programming, circuit design, and cross-platform migration across Arduino, ESP32, and Raspberry Pi Pico~\cite{xu2026embedagent}. IoT-SkillsBench evaluates programming agents across embedded platforms and peripherals through real hardware execution~\cite{li2026skilled}. EmbedCopilot-Bench extends evaluation to multimodal assistance for hardware operation, software configuration, and code generation~\cite{yuan2026embedcopilot}. These studies examine whether models can produce working implementations or useful development instructions. They do not provide a dedicated evaluation of constraint-grounded safety verdicts, including abstention when essential information is missing. BoardSafety addresses this gap by evaluating proposed operations against the documented constraints of the specified board.

\section{BoardSafety Benchmark}

\begin{figure*}[h]
    \centering
    \includegraphics[width=\textwidth]{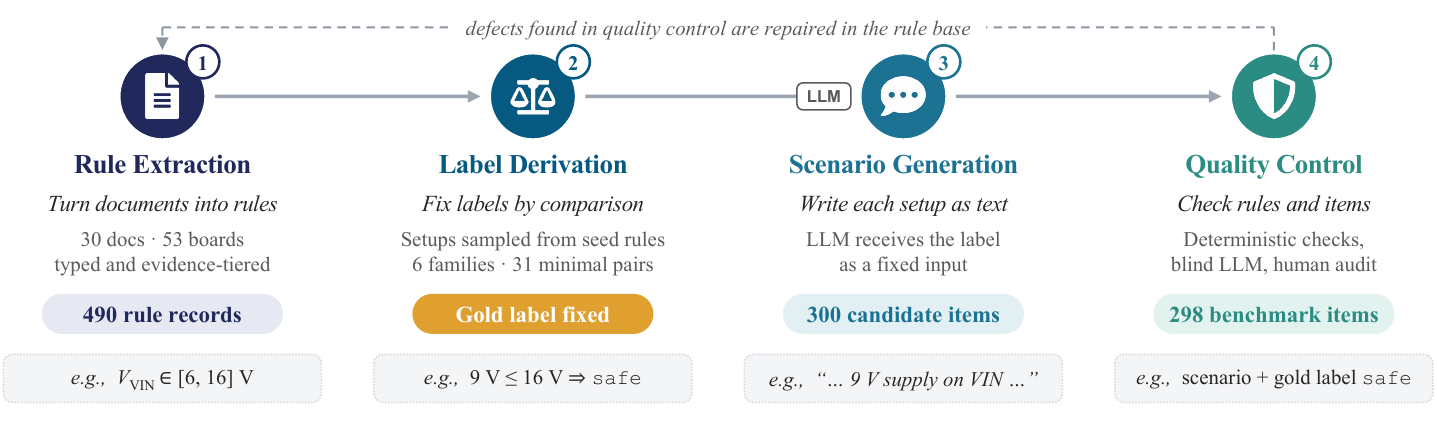}
    \caption{BoardSafety construction pipeline. We first extract board-specific constraints from manufacturer documentation and derive safety labels through rule-based comparisons. We then use LLMs to generate scenarios while preserving the fixed operating conditions and labels. Finally, we validate the rules and scenarios through deterministic checks, blind model validation, and human review.}
    \vspace{-2mm}
    \label{fig:pipeline}
    \vspace{-2mm}
\end{figure*}


BoardSafety evaluates whether an LLM can judge a proposed operation against the documented constraints of a specific development board. Each item specifies a board and an operating setup, and the model returns one of three verdicts: \texttt{safe}, \texttt{unsafe}, \texttt{insufficient-info}. Labels describe compliance with the selected vendor constraints under the stated scenario assumptions. The reference verdict is \texttt{unsafe} when the setup violates an applicable documented safety constraint, \texttt{insufficient-info} when no violation is established but information needed to determine compliance is missing, and \texttt{safe} when the available information establishes compliance with these constraints.

Figure~\ref{fig:pipeline} summarizes the construction pipeline. We first extract board-specific rules from vendor documentation and derive a reference label for each sampled operating setup. An LLM then expresses the setup as a natural-language scenario while preserving the fixed conditions and label. The resulting items undergo deterministic validation and additional model or human review.

\subsection{AI-Readiness}
Vendor documentation does not directly provide items that are ready for evaluation. because constraints are scattered across documents, mixed with descriptive specifications, and expressed in different forms. For LLM evaluation, we define AI-readiness using five properties that associate source evidence, structured constraints, and reference labels:
\smallskip
\begin{enumerate}[leftmargin=*,nosep]
    \item \textbf{Traceable}: selected rules retain source passages and document references for verification.

    \item \textbf{Computable}: numeric limits encode quantities, operators, values, and units for automatic comparison.

    \item \textbf{Filterable}: statement type and evidence tier determine which rules can seed benchmark items.

    \item \textbf{Consistent}: duplicate and conflicting bounds are resolved by identifying the applicable binding constraint.
    
    \item \textbf{Derivable}: each item records the comparison or configurational rule used to derive its label.
\end{enumerate}

\subsection{Source Corpus and Rule Base} 
We construct a rule base of 490 records across 17 hazard categories from 30 vendor documents covering 53 development boards from five vendors. We use GPT-5.6 Sol\cite{openai2026gpt56} to extract candidate rule records from vendor documentation and manually review a subset against the source passages. Each record stores a normalized constraint and source-document metadata. Of the 490 records, 487 also include a verbatim supporting passage.
Each record has a statement type and an evidence tier for rule selection. The statement type separates user-violable limits from descriptive specifications, internal-node properties, and procedural requirements. Evidence tiers distinguish documented facts from inference. Tier A records state both the constraint and its numeric limit. Tier B records state a constraint without a numeric limit. In Tier C1, the limit is documented but the damage pathway is inferred. In Tier C2, both are inferred.

\subsection{Rule Selection and Label Derivation}

The standard seed pool contains 304 Tier A/B records classified as user-violable design limits. At sampling time, we exclude the 17 records concerning cumulative effects because our scenario representation does not model exposure duration, leaving 287 records in this pool. For the underspecified family, we additionally allow Tier C1 records that quote a limit but infer the damage pathway, because these items test whether the available information is sufficient to evaluate the stated constraint. Family-level quotas, per-board caps, and a coverage sweep yield 300 candidate items from 213 seed rules.

For an eligible rule $r$ and sampling regime $g$, we define
\begin{equation}
  \sigma(r, g) \rightarrow (x, \ell, a),
\end{equation}
where $x$ is an operating condition, $\ell$ is its gold label with respect to the selected constraint, and $a$ is the rule-based justification for that label.

For the scenario families listed in Table~\ref{tab:families}, the sampling regimes specify how operating scenarios and labels are constructed. Violation regimes produce \texttt{unsafe} items by placing the relevant quantity beyond the selected documented limit or specifying a prohibited configuration. Two regimes produce \texttt{safe} items: in-spec items stay at least 15\% below a single upper limit, or at least 15\% of the interval width from each endpoint of a bounded range, whereas documented-safe items describe an operation explicitly permitted by the vendor under the stated conditions. Boundary pairs place the relevant quantity at 95\% and 105\% of the same upper limit, producing opposite labels. Board-swap pairs hold the operating setup fixed while changing the board. The selected value satisfies one board's limit and violates the other's, with a margin of at least 10\% relative to each upper limit; rule pairs without a feasible shared value are skipped. The same setup is therefore labeled \texttt{safe} for one board and \texttt{unsafe} for the other. The board-swap and boundary families contain 16 and 15 minimal pairs, respectively. Underspecified items omit a quantity required to evaluate the selected constraint. The remaining information must admit both satisfying and violating assignments to that quantity, so these items receive the label \texttt{insufficient-info}.

\begin{table}[t]
\centering
\caption{Scenario families and label distribution.}
\label{tab:families}
\small
\renewcommand{\arraystretch}{1.12}
\begin{tabular*}{\columnwidth}{@{}l@{\extracolsep{\fill}}lr@{}}
\toprule
Scenario family & Gold label & $n$ \\
\midrule
V: Constraint violation & \texttt{unsafe} & 113 \\
S: In-spec operation & \texttt{safe} & 58 \\
U: Underspecified setup & \texttt{insufficient-info} & 41 \\
B: Board swap & \texttt{safe} / \texttt{unsafe} & 32 \\
N: Near-boundary setup & \texttt{safe} / \texttt{unsafe} & 30 \\
D: Documented-safe operation & \texttt{safe} & 24 \\
\midrule
Total & & 298 \\
\bottomrule
\end{tabular*}
\vspace{-3mm}
\end{table}

\subsection{Rule-Grounded Scenario Generation}

After fixing $x$ and $\ell$, we use Claude Opus~5~\cite{anthropic2026claudeopus5} to render each rule instance as a natural-language scenario:
\begin{equation}
\rho(r,g,x) \rightarrow s,
\end{equation}
It takes the seed rule and the complete rule set of the board, with $x$ and $\ell$ as immutable inputs. The generator also returns an \texttt{asserted\_facts} record that links each stated quantity to a specific span in $s$. This enables the validator to verify that $x$ is stated correctly in $s$. Combined quantities, such as the total current of four LEDs drawing 10\,mA each from one pin, must be explicitly stated when they determine the label, rather than left to the reader to deduce. Distractors, which are quantities stated other than the decisive one, must adhere to all rules of the board. If a distractor conflicts with these rules, it is revised without altering $x$ or $\ell$, and the item is re-validated. In the case of \texttt{insufficient-info} items, the decisive quantity is intentionally omitted, but all stated quantities remain consistent with the rule set. Furthermore, both items of a minimal pair are realized from a single narrative and differ in the board identity or decisive value, with any accompanying changes restricted to details required by that change.

\subsection{Escalating Quality Control}

Quality control combines automatic checks, blind model validation, and human review. Of the 298 retained items, blind validation covers the pre-repair versions of 218 items, and human audit covers 98, with 91 items in both groups. These reviews cover 225 items, and all retained items pass deterministic validation.

\textbf{Provenance audit.} To verify extracted records against their sources, two reviewers independently examine a stratified sample of 59 records. They confirm 55 records and correct 4. All 59 records are traceable to supporting passages.

\textbf{Deterministic validation.} To check that generated scenarios preserve the fixed conditions and labels, a rule-based validator checks stated values, structured facts, and label consistency. All 300 candidate items pass these checks. Because validation relies on the extracted records, it cannot detect errors in their interpretation of the source documents.

\textbf{Blind validation.} To identify errors that deterministic validation misses, GPT-5.6~\cite{openai2026gpt56} independently labels 220 items without access to gold labels or metadata, agreeing on 209. Reviewing the complete board rule set leads to repairs to 31 rule records. After regeneration, a 60-item re-run agrees with the derived label on 56.

\textbf{Human audit.} To address cases that cannot be settled by the automated stages, two auditors with embedded-system experience independently review 100 items and reconcile any disagreements. Among the 11 disagreements between blind validation and the derived labels, they retain 9 labels and exclude 2 items whose seed records do not express user-violable constraints.

\subsection{Released Benchmark}

The released benchmark consists of 298 items across 46 of the boards, along with 211 seed rules. It contains 113 \texttt{safe}, 144 \texttt{unsafe}, and 41 \texttt{insufficient-info} items. Each board contributes between 1 and 16 items, so we provide aggregate and family-level results rather than rankings for each board. Each item is accompanied by its seed rule, the verbatim source passage, the evidence tier, the comparison $a$ from which its label is derived, and its validation and audit outcomes. This detailed information allows each label to be traced back to a vendor sentence and recomputed without a model.

\begin{table*}[t]
\centering
\caption{Accuracy (\%) across rule-access and rationale-format conditions. $T_0$ uses direct-answer prompting; $T_1$ adds a four-step structured rationale.  $\Delta=T_1-T_0$ in percentage points. Pooled values average $R_0$ and $R_1$.}
\label{tab:accuracy}
\small
\renewcommand{\arraystretch}{1.12}
\begin{tabular*}{\textwidth}{@{\extracolsep{\fill}}lccc ccc ccc@{}}
\toprule
& \multicolumn{3}{c}{$R_0$: No Rule Context} & \multicolumn{3}{c}{$R_1$: Explicit Rule Context} & \multicolumn{3}{c}{Pooled over $R_0,R_1$} \\
\cmidrule(lr){2-4}\cmidrule(lr){5-7}\cmidrule(lr){8-10}
Model & $T_0$ & $T_1$ & $\Delta$ & $T_0$ & $T_1$ & $\Delta$ & $T_0$ & $T_1$ & $\Delta$ \\
\midrule
Gemma-3-12B  & 47.32 & 52.01 & +4.70 & 60.07 & 69.13 & +9.06 & 53.69 & 60.57 & +6.88 \\
Gemma-3-1B   & 37.58 & 47.65 & +10.07 & 45.30 & 46.31 & +1.01 & 41.44 & 46.98 & +5.54 \\
Llama-3.1-8B & 45.97 & 49.66 & +3.69 & 55.03 & 53.02 & -2.01 & 50.50 & 51.34 & +0.84 \\
Llama-3.2-3B & 36.91 & 44.63 & +7.72 & 49.33 & 49.33 & 0.00 & 43.12 & 46.98 & +3.86 \\
Qwen3-0.6B   & 36.91 & 40.60 & +3.69 & 45.97 & 50.67 & +4.70 & 41.44 & 45.64 & +4.19 \\
Qwen3-4B     & 40.27 & 37.58 & -2.68 & 52.68 & 65.77 & +13.09 & 46.48 & 51.68 & +5.20 \\
\midrule
\textbf{Mean} & \textbf{40.83} & \textbf{45.36} & \textbf{+4.53} & \textbf{51.40} & \textbf{55.70} & \textbf{+4.31} & \textbf{46.11} & \textbf{50.53} & \textbf{+4.42} \\
\bottomrule
\end{tabular*}
\vspace{-2mm}
\end{table*}

\section{Experiments}
\label{sec:experiments}

\subsection{Experimental Setting}

\textbf{Implementation details. }We evaluate six open-weight instruction-tuned LLMs on all 298 BoardSafety scenarios across 46 development boards: Qwen3-0.6B and Qwen3-4B~\cite{yang2025qwen3}, Gemma-3-1B and Gemma-3-12B~\cite{team2025gemma}, and Llama-3.2-3B-Instruct and Llama-3.1-8B-Instruct~\cite{grattafiori2024llama}. These models cover three families, with a smaller and a larger model from each family. All conditions use temperature 0 and a generation cap of 2,048 tokens, with native thinking disabled for Qwen3~\cite{yang2025qwen3}. Only formatting failures trigger retries. We score the final valid response for each item.

We use a $2\times2$ design that varies rule access and rationale prompting. Under $R_0$, models receive only the board name and scenario. Under $R_1$, they additionally receive normalized constraints and limits for that board. Under $T_0$, models return a verdict with a brief justification of one to two sentences. Under $T_1$, they additionally provide four rationale steps covering the relevant constraint, decisive quantity, comparison, and conclusion, with a specified length limit for each step. Each model answers every scenario under $R_0T_0$, $R_0T_1$, $R_1T_0$, and $R_1T_1$, producing 7,152 responses.


\textbf{Metrics. }We define accuracy as the proportion of verdicts matching the gold labels and report 95\% Wilson confidence intervals. The always-unsafe baseline predicts \texttt{unsafe} for every item and achieves $144/298=48.3\%$ accuracy. Dangerous acceptance refers to the proportion of \texttt{unsafe} items predicted as \texttt{safe}, and safe rejection indicates the proportion of \texttt{safe} items predicted as \texttt{unsafe}. We also examine accuracy by scenario family, the frequency of \texttt{insufficient-info} predictions, and the proportion of board-swap and boundary pairs for which both items are correctly labeled.

\textbf{Explanation. }We evaluate model rationales using rationale length and the correctness of constraint application. After the model generates its response, we measure the length of the brief justification under $T_0$ and the four-step rationale under $T_1$ using the model's tokenizer. To assess correctness, we sample three responses per model, condition, and gold label, for a total of 216 responses. A single LLM judge evaluates each response using the original prompt, reference rule, and reference comparison, with model identity withheld. It checks rule selection, the decisive quantity, the comparison, and the conclusion. A response passes only if all four are correct. We report the percentage of sampled responses that pass.

\IfFileExists{figures/fig3_error.pdf}{
\begin{figure*}[h]
    \centering
    \includegraphics[width=\textwidth]{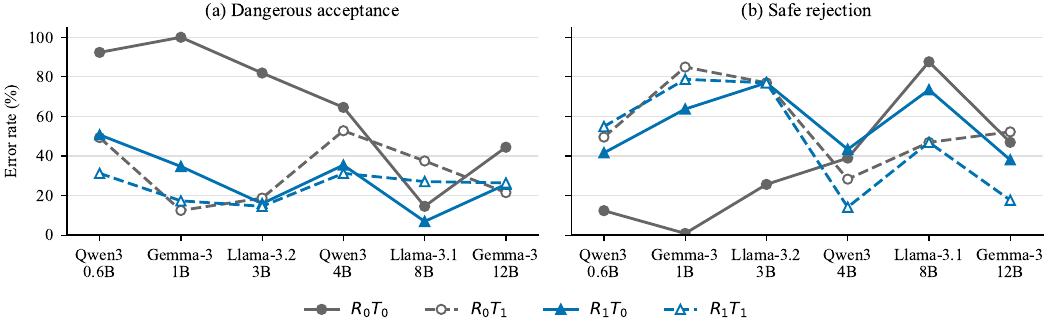}
    \caption{Safety judgment errors across rule-access and rationale-prompting conditions. Panel (a) shows the proportion of unsafe operations classified as safe; panel (b) shows the proportion of safe operations classified as unsafe. Lower rates indicate better performance.}
    \label{fig:error-asymmetry}
    \vspace{-2mm}
\end{figure*}
\vspace{-2mm}
}{}

\subsection{RQ1: How Do Rule Access and Rationale Prompting Affect the Reliability of Safety Judgments?}

To evaluate judgment reliability, we compare accuracy against the always-unsafe baseline and examine dangerous acceptance and safe rejection. We compare rule access at each fixed rationale-prompting condition and rationale prompting at each fixed rule-access condition. Table~\ref{tab:accuracy} reports accuracy across all model--condition combinations, ranging from 36.91\% to 69.13\%. Gemma-3-12B achieves the highest accuracy under $R_1T_1$, compared with the 48.3\% baseline. Only 4 of the 24 combinations have a 95\% Wilson lower bound that exceeds this baseline.

Rule access produces the larger average accuracy gain, as shown in Table~\ref{tab:accuracy}. On average, it increases accuracy by 10.57 percentage points under $T_0$ and 10.35 points under $T_1$. Rationale prompting increases mean accuracy by 4.53 percentage points without rules and 4.31 points with rules. Its benefit varies across models and rule-access conditions. For Qwen3-4B, the change is $-2.68$ points without rules and $+13.09$ points with rules. These results show that rule access provides the larger average improvement, while structured-rationale prompting does not improve every model.

Figure~\ref{fig:error-asymmetry} shows that substantial safety errors remain despite the accuracy gains. On average, dangerous acceptance falls from 66.3\% under $R_0T_0$ to 24.7\% under $R_1T_1$, while safe rejection rises from 35.4\% to 48.2\%. This comparison reflects the combined changes in rule access and rationale prompting. Fewer unsafe operations receive approval, but more safe operations are rejected. Thus, higher accuracy does not ensure reliable safety judgments.

\subsection{RQ2: What Does Aggregate Accuracy Obscure?}

To examine what aggregate accuracy may obscure, we compare performance across scenario families and check if both items of each minimal pair receive the correct labels. Table~\ref{tab:family-accuracy} shows the family-level spread under $R_1T_1$: all six models score only 0--19.5\% on underspecified cases, versus 66.4--84.1\% on violation cases. Despite its 69.13\% overall accuracy, Gemma-3-12B reaches 86\% on in-spec cases and 77\% on violation cases but only 63\% on boundary cases. It correctly labels both items in 11 of 16 board-swap pairs and 4 of 15 boundary pairs. Therefore, the overall accuracy does not demonstrate that the model consistently distinguishes changes in board-specific limits or operating values near a boundary.

\begin{table}[!htbp]
\centering
\caption{Per-item accuracy (\%) by scenario family and model under $R_1T_1$.}
\label{tab:family-accuracy}
\footnotesize
\setlength{\tabcolsep}{1.5pt}
\renewcommand{\arraystretch}{1.2}
\begin{tabular*}{\columnwidth}{@{}l@{\extracolsep{\fill}}rrrrrr@{}}
\toprule
& \multicolumn{2}{c}{Gemma-3} & \multicolumn{2}{c}{Llama} & \multicolumn{2}{c}{Qwen3} \\
\cmidrule(lr){2-3}\cmidrule(lr){4-5}\cmidrule(l){6-7}
Family & 1B & 12B & 3.2-3B & 3.1-8B & 0.6B & 4B \\
\midrule
In-spec & 20.7 & 86.2 & 25.9 & 55.2 & 51.7 & 91.4 \\
Violation & 80.5 & 77.0 & 84.1 & 71.7 & 70.8 & 66.4 \\
Doc. safe & 16.7 & 66.7 & 29.2 & 58.3 & 25.0 & 79.2 \\
Boundary & 46.7 & 63.3 & 46.7 & 50.0 & 53.3 & 66.7 \\
Board swap & 53.1 & 81.3 & 50.0 & 43.8 & 56.3 & 71.9 \\
Underspec. & 0.0 & 19.5 & 0.0 & 4.9 & 2.4 & 14.6 \\
\midrule
Overall & 46.3 & 69.1 & 49.3 & 53.0 & 50.7 & 65.8 \\
\bottomrule
\end{tabular*}
\end{table}

We also notice that the models struggle to recognize missing information. Under $R_1T_1$, they predict \texttt{insufficient-\allowbreak{}info} for only 0.3--4.0\% of all scenarios, although underspecified cases account for 13.8\% of the benchmark. Gemma-3-12B correctly identifies only 8 of the 41 underspecified cases, yielding an accuracy of 19.5\% on this subset. Moreover, 26 of the 27 scenarios answered incorrectly by all six models under this condition are underspecified. These results show that models frequently give definite verdicts despite lacking the information needed to assess safety. Together with the board-swap and near-boundary results, this finding highlights limitations that aggregate accuracy alone does not capture. Safety evaluation should therefore assess sensitivity to board-specific limits, handling of operating boundaries, and abstention when essential information is missing.

\subsection{RQ3: Do LLM Responses Follow the Safety Constraints?}

To examine whether model rationales effectively support their verdicts, we evaluate a sample of responses using four criteria for applying constraints. Table~\ref{tab:rq3-criteria} reports the pass rates for each criterion and for all four jointly. Only 13.4\% of these responses pass all four criteria. Among the responses that provide a correct verdict, 62.8\% fail to satisfy at least one criterion. These findings show that a correct verdict does not necessarily come with a well-supported explanation. The reported rates describe the stratified review sample, which gives equal weight to the three gold labels.

\begin{table}[!b]
\centering
\caption{Constraint-application pass rates in the stratified review (216 responses and 54 per condition).}
\label{tab:rq3-criteria}
\small
\renewcommand{\arraystretch}{1.2}
\begin{tabular*}{\columnwidth}{@{}l@{\extracolsep{\fill}}rr@{}}
\toprule
Criterion & Correct / reviewed & Pass rate \\
\midrule
Rule selection & 47/216 & 21.8\% \\
Decisive quantity & 63/216 & 29.2\% \\
Comparison / logical test & 31/216 & 14.4\% \\
Conclusion & 78/216 & 36.1\% \\
\addlinespace[2pt]
\textbf{All four correct} & \textbf{29/216} & \textbf{13.4\%} \\
\midrule
\addlinespace[2pt]
Condition (all four) & Correct / reviewed & Pass rate \\
\midrule
$R_0T_0$ & 3/54 & 5.6\% \\
$R_0T_1$ & 6/54 & 11.1\% \\
$R_1T_0$ & 13/54 & 24.1\% \\
$R_1T_1$ & 7/54 & 13.0\% \\
\bottomrule
\end{tabular*}
\end{table}

We examine whether structured-rationale prompting improves constraint application. Across all 7,152 responses, the median visible-rationale length increases from 32 tokens under $T_0$ to 79 under $T_1$. However, Table~\ref{tab:rq3-criteria} shows that, in the reviewed sample with rule access, constraint-application correctness is 24.1\% under $R_1T_0$ and 13.0\% under $R_1T_1$. These results show that structured-rationale prompting does not necessarily improve constraint application, even when it produces longer explanations. Since the prompting conditions also differ in their required reasoning steps, the observed difference cannot be attributed to rationale length alone. These findings highlight the need to evaluate constraint application directly, as longer explanations do not necessarily indicate correct application.

\section{Conclusion and Future Work}

Safe hardware advice requires LLMs to follow board-specific operating limits and recognize when essential information is missing. To evaluate these capabilities, we introduce BoardSafety, a benchmark for judging whether proposed development-board operations comply with manufacturer-documented constraints. We build the benchmark from vendor datasheets and hardware guides, extracting structured constraints and using them to derive reference labels. We then generate operating scenarios with fixed labels and validate them through deterministic checks, blind model validation, and human review.
We conduct several experiments on BoardSafety to examine how rule access and structured-rationale prompting affect LLM safety judgments. Our experiments show that providing relevant rules improves performance but does not ensure correct constraint application. Models still struggle with board-specific limits, operating boundaries, and missing information. Correct verdicts can also be supported by incorrect rationales. These findings show that reliable safety evaluation requires checking constraint application and abstention alongside verdict accuracy.

Our future work would extend the evaluation to more models and practical hardware scenarios, strengthen expert validation, and test retrieval combined with deterministic constraint checks.

\section*{Acknowledgments}
This work is supported in part by the National Science Foundation (NSF) under grants 2242812, 2550105, 2550106, 2543905, and 2616468.

\bibliographystyle{IEEEtran}
\bibliography{refs}

@inproceedings{levy2022safetext,
  title={SafeText: A benchmark for exploring physical safety in language models},
  author={Levy, Sharon and Allaway, Emily and Subbiah, Melanie and Chilton, Lydia and Patton, Desmond and McKeown, Kathleen and Wang, William Yang},
  booktitle={Proceedings of the 2022 Conference on Empirical Methods in Natural Language Processing},
  pages={2407--2421},
  year={2022}
}

@article{hager2024evaluation,
  title={Evaluation and mitigation of the limitations of large language models in clinical decision-making},
  author={Hager, Paul and Jungmann, Friederike and Holland, Robbie and Bhagat, Kunal and Hubrecht, Inga and Knauer, Manuel and Vielhauer, Jakob and Makowski, Marcus and Braren, Rickmer and Kaissis, Georgios and others},
  journal={Nature medicine},
  volume={30},
  number={9},
  pages={2613--2622},
  year={2024},
  publisher={Nature Publishing Group US New York}
}

@article{zhou2024labsafety,
  title={Labsafety bench: Benchmarking llms on safety issues in scientific labs},
  author={Zhou, Yujun and Yang, Jingdong and Huang, Yue and Guo, Kehan and Emory, Zoe and Ghosh, Bikram and Bedar, Amita and Shekar, Sujay and Liang, Zhenwen and Chen, Pin-Yu and others},
  journal={arXiv preprint arXiv:2410.14182},
  year={2024}
}

@article{wu2025first,
  title={First, do NOHARM: a medical safety benchmark and randomized study of physician and AI teaming on clinical consultations},
  author={Wu, David and Haredasht, Fateme Nateghi and Maharaj, Saloni Kumar and Jain, Priyank and Tran, Jessica and Gwiazdon, Matthew and Rustagi, Arjun and Jindal, Jenelle and Koshy, Jacob M and Kadiyala, Vinay and others},
  journal={arXiv preprint arXiv:2512.01241},
  year={2025}
}

@article{skelic2025circuit,
  title={CIRCUIT: A benchmark for circuit interpretation and reasoning capabilities of LLMs},
  author={Skelic, Lejla and Xu, Yan and Cox, Matthew and Lu, Wenjie and Yu, Tao and Han, Ruonan},
  journal={arXiv preprint arXiv:2502.07980},
  year={2025}
}

@article{srinivasan2026pcb,
  title={PCB-QA: Evaluating LLMs over the First Printed Circuit Board Design Question-Answer Dataset},
  author={Srinivasan, Sahana and Tan, Benjamin and Turnbull, Benjamin and Pearce, Hammond},
  journal={arXiv preprint arXiv:2606.23704},
  year={2026}
}

@inproceedings{li2026pcb,
  title={Pcb-bench: Benchmarking llms for printed circuit board placement and routing},
  author={Li, Jindong and Chen, Lianrong and Yang, Bin and Zhu, Jiadong and Wang, Ying and Ma, Yuzhe and Yang, Menglin},
  booktitle={The Fourteenth International Conference on Learning Representations},
  year={2026}
}

@article{qiu2026hwe,
  title={HWE-Bench: Can Language Models Perform Board-level Schematic Designs?},
  author={Qiu, Weibo and Xiao, Yinhao and Pan, Runyu},
  journal={arXiv preprint arXiv:2603.18102},
  year={2026}
}

@inproceedings{englhardt2024exploring,
  title={Exploring and characterizing large language models for embedded system development and debugging},
  author={Englhardt, Zachary and Li, Richard and Nissanka, Dilini and Zhang, Zhihan and Narayanswamy, Girish and Breda, Joseph and Liu, Xin and Patel, Shwetak and Iyer, Vikram},
  booktitle={Extended Abstracts of the CHI Conference on Human Factors in Computing Systems},
  pages={1--9},
  year={2024}
}

@inproceedings{xu2026embedagent,
  title={Embedagent: Benchmarking large language models in embedded system development},
  author={Xu, Ruiyang and Cao, Jialun and Wu, Mingyuan and Zhong, Wenliang and Lu, Yaojie and He, Ben and Han, Xianpei and Cheung, Shing-Chi and Sun, Le},
  booktitle={Proceedings of the 2026 IEEE/ACM 48th International Conference on Software Engineering},
  pages={630--641},
  year={2026}
}

@article{yang2025qwen3,
  title={Qwen3 technical report},
  author={Yang, An and Li, Anfeng and Yang, Baosong and Zhang, Beichen and Hui, Binyuan and Zheng, Bo and Yu, Bowen and Gao, Chang and Huang, Chengen and Lv, Chenxu and others},
  journal={arXiv preprint arXiv:2505.09388},
  year={2025}
}

@inproceedings{yang2026thoughts,
  title={Do Thoughts Depth Affect Multilingual Reasoning?},
  author={Yang, Linjian and Wang, Xinyan and Liu, Kunpeng},
  booktitle={Proceedings of the 1st Workshop on Multilinguality in the Era of Large Language Models (MeLLM 2026)},
  pages={149--158},
  year={2026}
}

@article{team2025gemma,
  title={Gemma 3 technical report},
  author={Team, Gemma and Kamath, Aishwarya and Ferret, Johan and Pathak, Shreya and Vieillard, Nino and Merhej, Ramona and Perrin, Sarah and Matejovicova, Tatiana and Ram{\'e}, Alexandre and Rivi{\`e}re, Morgane and others},
  journal={arXiv preprint arXiv:2503.19786},
  year={2025}
}

@article{grattafiori2024llama,
  title={The llama 3 herd of models},
  author={Grattafiori, Aaron and Dubey, Abhimanyu and Jauhri, Abhinav and Pandey, Abhinav and Kadian, Abhishek and Al-Dahle, Ahmad and Letman, Aiesha and Mathur, Akhil and Schelten, Alan and Vaughan, Alex and others},
  journal={arXiv preprint arXiv:2407.21783},
  year={2024}
}

@inproceedings{xu2024knowledge,
  title={Knowledge conflicts for llms: A survey},
  author={Xu, Rongwu and Qi, Zehan and Guo, Zhijiang and Wang, Cunxiang and Wang, Hongru and Zhang, Yue and Xu, Wei},
  booktitle={Proceedings of the 2024 Conference on Empirical Methods in Natural Language Processing},
  pages={8541--8565},
  year={2024}
}

@inproceedings{xie2024adaptive,
  title={Adaptive chameleon or stubborn sloth: Revealing the behavior of large language models in knowledge conflicts},
  author={Xie, Jian and Zhang, Kai and Chen, Jiangjie and Lou, Renze and Su, Yu},
  booktitle={International Conference on Learning Representations},
  volume={2024},
  pages={35623--35646},
  year={2024}
}

@article{wang2023resolving,
  title={Resolving knowledge conflicts in large language models},
  author={Wang, Yike and Feng, Shangbin and Wang, Heng and Shi, Weijia and Balachandran, Vidhisha and He, Tianxing and Tsvetkov, Yulia},
  journal={arXiv preprint arXiv:2310.00935},
  year={2023}
}

@inproceedings{xu2026domusmind,
  title={DomusMind: A Benchmark for Evaluating Lifelong Smart Home Agents Under Drift},
  author={Xu, Rong and Wan, Yinxin and Xue, Xiaochan},
  booktitle={ICLR 2026 Workshop on Lifelong Agents: Learning, Aligning, Evolving},
  year={2026}
}

@article{chen2024doceda,
  title={DocEDA: Automated extraction and design of analog circuits from documents with large language model},
  author={Chen, Hong Cai and Wu, Longchang and Gao, Ming and Shen, Lingrui and Zhong, Jiarui and Xu, Yipin},
  journal={arXiv preprint arXiv:2412.05301},
  year={2024}
}

@article{wilkinson2016fair,
  title={The FAIR Guiding Principles for scientific data management and stewardship},
  author={Wilkinson, Mark D and Dumontier, Michel and Aalbersberg, IJsbrand Jan and Appleton, Gabrielle and Axton, Myles and Baak, Arie and Blomberg, Niklas and Boiten, Jan-Willem and da Silva Santos, Luiz Bonino and Bourne, Philip E and others},
  journal={Scientific data},
  volume={3},
  number={1},
  pages={160018},
  year={2016},
  publisher={Nature Publishing Group}
}

@article{ma2023comprehensive,
  title={A comprehensive survey on vector database: Storage and retrieval technique, challenge},
  author={Ma, Le and Zhang, Ran and Han, Yikun and Yu, Shirui and Wang, Zaitian and Ning, Zhiyuan and Zhang, Jinghan and Xu, Ping and Li, Pengjiang and Ju, Wei and others},
  journal={arXiv preprint arXiv:2310.11703},
  year={2023}
}

@inproceedings{madhusudhan2025llms,
  title={Do llms know when to not answer? investigating abstention abilities of large language models},
  author={Madhusudhan, Nishanth and Madhusudhan, Sathwik Tejaswi and Yadav, Vikas and Hashemi, Masoud},
  booktitle={Proceedings of the 31st International Conference on Computational Linguistics},
  pages={9329--9345},
  year={2025}
}

@inproceedings{li2026skilled,
  title={Skilled AI Agents for Embedded and IoT Systems Development},
  author={Li, Yiming and Cheng, Yuhan and Ma, Mingchen and Zou, Yihang and Yang, Ningyuan and Cheng, Wei and Li, Hai and Chen, Yiran and Chen, Tingjun},
  booktitle={Proceedings of the ACM Conference on AI and Agentic Systems},
  pages={1248--1252},
  year={2026}
}

@inproceedings{yuan2026embedcopilot,
  title={EmbedCopilot-Bench: Evaluating Vision-Language Models for Hardware-Aware Embedded System Development},
  author={Yuan, Dongsheng and Deng, Yimo and Chen, Huangxun},
  booktitle={European Conference on Computer Vision},
  pages={332--350},
  year={2026},
  organization={Springer}
}

@article{zhang2026braillebench,
  title={BrailleBench: Investigating Multi-Criteria Braille Comprehension in Large Language Models},
  author={Zhang, Jinghan and Mo, Fengran and Chen, Zhiyu and Han, Xiaoyan and Liu, Kunpeng and Lu, Chang-Tien},
  journal={arXiv preprint arXiv:2608.27268},
  year={2026}
}

@inproceedings{zhang2025ratt,
  title={Ratt: A thought structure for coherent and correct llm reasoning},
  author={Zhang, Jinghan and Wang, Xiting and Ren, Weijieying and Jiang, Lu and Wang, Dongjie and Liu, Kunpeng},
  booktitle={Proceedings of the AAAI Conference on Artificial Intelligence},
  volume={39},
  number={25},
  pages={26733--26741},
  year={2025}
}

@inproceedings{mo2026opendecoder,
  title={Opendecoder: Open large language model decoding to incorporate document quality in rag},
  author={Mo, Fengran and Su, Zhan and Hui, Yuchen and Zhang, Jinghan and Sun, Jia Ao and Liu, Zheyuan and Zhang, Chao and Sakai, Tetsuya and Nie, Jian-Yun},
  booktitle={Proceedings of the ACM Web Conference 2026},
  pages={2252--2262},
  year={2026}
}

@inproceedings{xu2026shift,
  title={Shift-Resilient Diffusive Imputation for Variable Subset Forecasting},
  author={Xu, Haihua and Hao, Qi and Zhang, He and Zhao, Jianpeng and Qiao, Ziyue and Jiang, Lu and Wang, Pengfei and Zhou, Yingjie and Wang, Pengyang},
  booktitle={Proceedings of the ACM Web Conference 2026},
  pages={7366--7377},
  year={2026}
}

@inproceedings{hao2025generative,
  title={Generative Imputation with Multi-level Causal Consistency for Variable Subset Forecasting},
  author={Hao, Qi and Gao, Yue and Liang, Runchang and Zhang, Yunhe and Wang, Pengyang},
  booktitle={Proceedings of the 31st ACM SIGKDD Conference on Knowledge Discovery and Data Mining V. 2},
  pages={838--849},
  year={2025}
}

@misc{anthropic2026claudeopus5,
  author       = {{Anthropic}},
  title        = {{Claude Opus 5} System Card},
  year         = {2026},
  howpublished = {\url{https://www.anthropic.com/claude-opus-5-system-card}},
  note         = {Accessed: 2026-09-09}
}

@misc{openai2026gpt56,
  author       = {{OpenAI}},
  title        = {{GPT-5.6} System Card},
  year         = {2026},
  howpublished = {\url{https://deploymentsafety.openai.com/gpt-5-6}},
  note         = {Accessed: 2026-09-09}
}

@misc{arduino_due_ds,
  author       = {{Arduino}},
  title        = {Arduino {Due} User Manual, {SKU} {A000062}},
  howpublished = {\url{https://docs.arduino.cc/resources/datasheets/A000062-datasheet.pdf}},
  year         = {2026},
  note         = {Accessed: 2026-09-09}
}

@misc{arduino_mkr1010_ds,
  author       = {{Arduino}},
  title        = {Arduino {MKR} {WiFi} 1010 User Manual, {SKU} {ABX00023}},
  howpublished = {\url{https://docs.arduino.cc/resources/datasheets/ABX00023-datasheet.pdf}},
  year         = {2026},
  note         = {Accessed: 2026-09-09}
}

\end{document}